\pdfoutput=1
\documentclass[10pt,twocolumn,letterpaper]{article}

\usepackage{cvpr}
\usepackage{times}
\usepackage{epsfig}
\usepackage{graphicx}
\usepackage{amsmath}
\usepackage{amssymb}
\usepackage{subcaption}

\usepackage[pagebackref=true,breaklinks=true,colorlinks,bookmarks=false]{hyperref}

\cvprfinalcopy 

\begin{document}

\title{Deep Residual Networks with Exponential Linear Unit}

\author{Anish Shah \qquad Sameer Shinde \qquad Eashan Kadam \qquad Hena Shah \qquad Sandip Shingade\\
Veermata Jijabai Technological Institute\\
Mumbai, India\\
}

\maketitle

\begin{abstract}
    Very deep convolutional neural networks introduced new 
    problems like \emph{vanishing gradient} and \emph{degradation}.
    The recent successful contributions towards solving these problems are Residual and Highway Networks.
    These networks introduce \emph{skip connections} that allow the information (from the input or those learned in earlier layers) to flow more into the deeper layers.
    These very deep models have lead to a considerable decrease in test errors, on benchmarks like ImageNet and COCO.
    In this paper, we propose the use of \emph{exponential linear unit} instead of the combination of ReLU and Batch Normalization in Residual Networks.
    We show that this not only speeds up learning in Residual Networks but also improves the accuracy as the depth increases. 
    It improves the test error on almost all data sets, like CIFAR-10 and CIFAR-100.
\end{abstract}

\section{Introduction}

The Vision Community has been mesmerized by the effectiveness of deep convolutional neural networks (CNNs) \cite{[13]} that have led to a breakthrough in computer vision-related problems. Hence, there has been a notable shift towards CNNs in many areas of computer vision \cite{[12], [14], [15], [16], [17]}. Convolutional neural networks were popularized through AlexNet \cite{[10]} in 2009 and their much celebrated victory at the 2012 ImageNet competiton \cite{[11],[12]}. After that, there have been several attempts at building deeper and deeper CNNs like the VGG network and GoogLeNet in 2014 which have 19 and 22 layers respectively \cite{[15], [17]}. But, very deep models introduce problems like vanishing and exploding gradients \cite{[3]}, which hamper their convergence. 

The \emph{vanishing gradient} problem is trivial in very deep networks. During the backpropagation phase, the gradients are computed by the chain rule. Multiplication of small numbers in the chain rule leads to an exponential decrease in the gradient. Due to this, very deep networks learn very slowly. Sometimes, the gradient in the earlier layer gets larger because derivatives of some activation functions can take larger values. This leads to the problem of \emph{exploding gradient}. These problems have been reduced in practice through normalized initialization \cite{[3]} and most recently, Batch Normalization \cite{[4]}.

\emph{Exponential linear unit} (ELU) \cite{[9]} also reduces the vanishing gradient problem. ELUs introduce negative values which push the mean activation towards zero. This reduces the \emph{bias shift} and speeds up learning. ELUs give better accuracy and learning speed-up compared to the combination of ReLU \cite{[8]} and Batch Normalization \cite{[4]}.

After reducing the vanishing/exploding gradient problem, the networks start converging. However, the accuracy degrades in such very deep models \cite{[1]}. The most recent contributions towards solving this problem are Highway Networks \cite{[7]} and Residual Networks \cite{[1]}. These networks introduce \emph{skip connections}, which allow information flow into the deeper layers and enable us to have deeper networks with better accuracy. The 152-layer ResNet outperforms all other models \cite{[1]}. 

In this paper, we propose to use exponential linear unit instead of the combination of ReLU and Batch Normalization. Since exponential linear units reduce the vanishing gradient problem and give better accuracy compared to the combination of ReLU and Batch Normalization, we use it in our model to further increase the accuracy of Residual Networks. We also notice that ELU speeds up learning in very deep networks as well. We show that our model increases the accuracy on datasets like CIFAR-10 and CIFAR-100, compared to the original model. It is seen that as the depth increases, the difference in accuracy between our model and the original model increases.

\section{Background}

Deeper neural networks are very difficult to train. The vanishing/exploding gradients problem impedes the convergence of deeper networks \cite{[3]}. This problem has been solved by normalized initialization \cite{[3], [5], [6]}. A notable recent contribution towards reducing the vanishing gradients problem is Batch Normalization \cite{[4]}. Instead of normalized initialization and keeping a lower learning rate, Batch Normalization makes normalization a part of the model and performs it for each mini-batch. 

Once the deeper networks start converging, a \emph{degradation} problem occurs. Due to this, the accuracy degrades rapidly after it is saturated. The \emph{training error} increases as we add more layers to a deep model, as mentioned in \cite{[2]}. To solve this problem, several authors introduced skip connections to improve the information flow across several layers. Highway Networks \cite{[7]} have parameterized skip connections, known as \emph{information highways}, which allow information to flow unimpeded into deeper layers. During the training phase, the skip connection parameters are adjusted to control the amount of information allowed on these \emph{highways}.

\begin{figure}
\centering
\includegraphics[height=5cm]{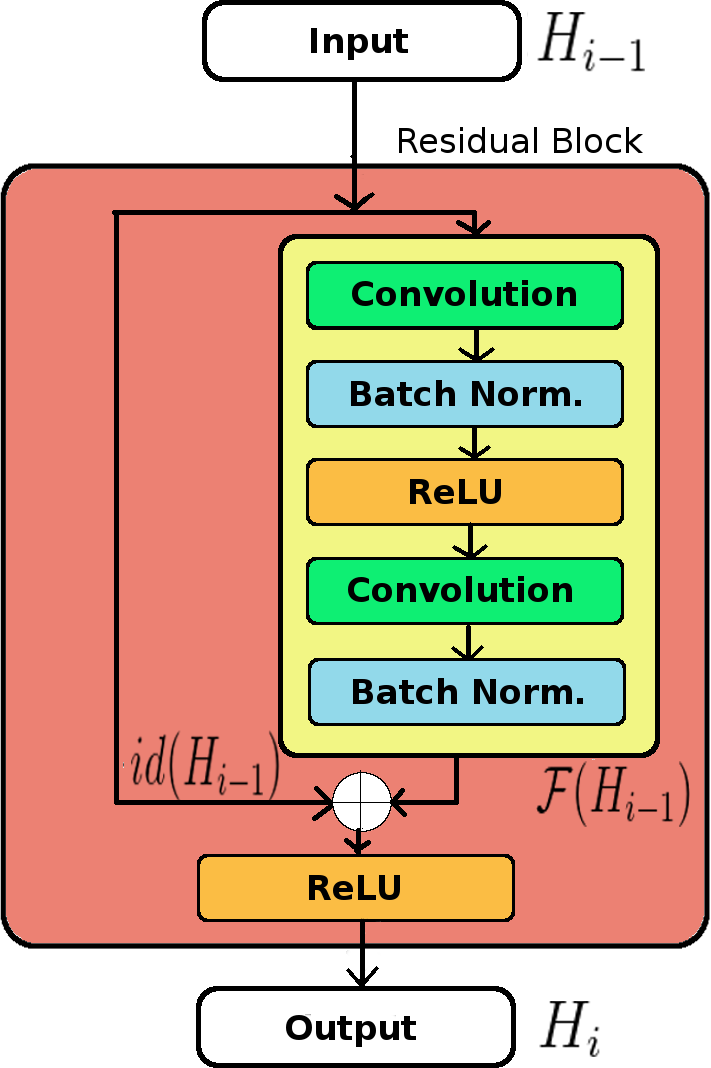}
\caption{An \( i^{th} \) Residual Block in a Residual Networks}
\label{fig:resblock}
\end{figure}

\textbf{Residual Networks (ResNets)} \cite{[1]} utilize shortcut connections with the help of identity transformation. Unlike Highway Networks, these neither introduce extra parameter nor computation complexity. This improves the accuracy of deeper networks. With increasing depth, ResNets give better function approximation capabilities as they gain more parameters. The authors' hypothesis is that the plain deeper networks give worse function approximation because the gradients vanish when they are propagated through many layers. To fix this problem, they introduce skip connections to the network. Formally, If the output of \( i^{th} \) layer is \( H_i \) and \( \mathcal{F} \) represents multiple convolutional transformation from layer \( i-1 \) to \( i \), we obtain

\begin{equation} \label{eq:1}
 H_i = ReLU( \mathcal{F}(H_{i-1}) + id(H_{i-1} ))
\end{equation}

where \( id(\cdot) \) represents the identity function and \( ReLU \) \cite{[8]} is the default activation function. Fig. \ref{fig:resblock} illustrates the basic building block of a Residual Network which consists of multiple convolutional and Batch Normalization layers. The identity transformation, \( id(\cdot) \) is used to reduce the dimensions of \( H_{i-1} \) to match those of \( \mathcal{F}(H_{i-1}) \). In Residual Networks, the gradients and features learned in earlier layers are passed back and forth between the layers via the identity transformations \( id(\cdot) \).

\textbf{Exponential Linear Unit (ELU)} \cite{[9]} alleviates the vanishing gradient problem and also speeds up learning in deep neural networks which leads to higher classification accuracies. The \emph{exponential linear unit} (ELU) is
\[ f(x) = \bigg \{ \begin{tabular}{cc}
$x$ & if $x > 0$ \\
$\alpha (exp(x)-1)$ & if $x \leq 0$
\end{tabular} \]

\begin{figure}
\centering
\includegraphics[height=5cm]{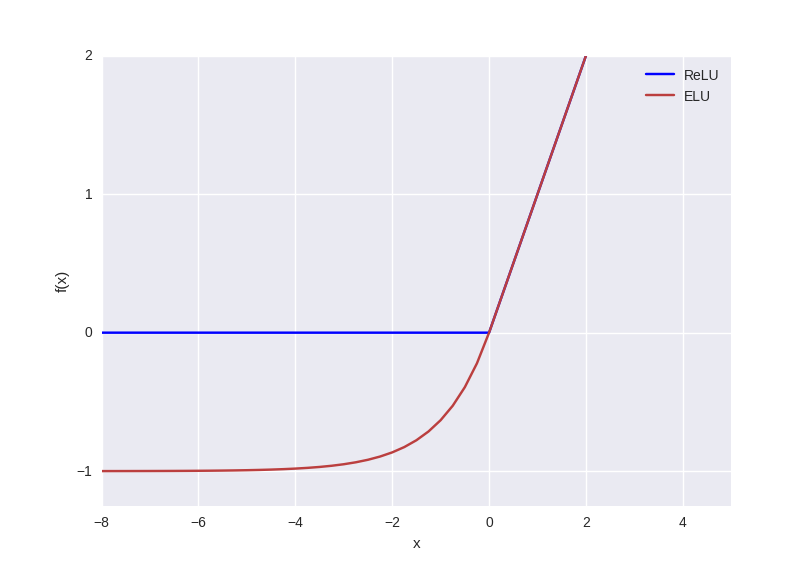}
\caption{The rectified linear unit (ReLU) and Exponential Linear Unit (ELU, $\alpha = 1.0$)}
\label{fig:elu}
\end{figure}

The ReLUs are non-negative and thus have mean activations larger than zero, whereas ELUs have negative values, which push the mean activations towards zero. ELUs saturate to a negative value when the input gets smaller. This decreases the forward propagated variation and information, which draws the mean activations to zero. Units with non-zero mean activations act as a bias for the next layer. If these units do not cancel each other out, then the learning causes a \emph{bias shift} for units in the next layer. Therefore, ELUs decrease the bias shift as the mean activations are closer to zero. Less bias shift also speeds up learning by bringing standard gradient closer towards the unit natural gradient. Fig.~\ref{fig:elu} shows the comparison of ReLU and ELU ($\alpha=1.0$).

\begin{figure*}
    \centering
    \begin{subfigure}{.24\linewidth}
        \centering
        \includegraphics[height=5cm]{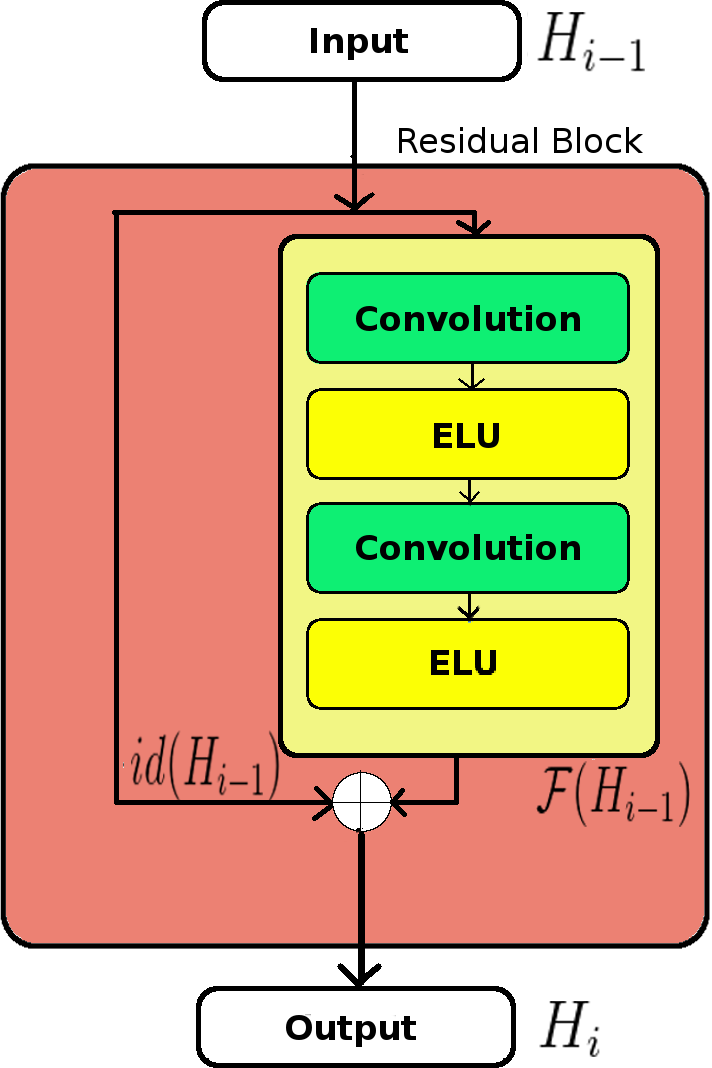}
        \caption{Conv-ELU-Conv-ELU}
        \label{fig:Conv-ELU-Conv-ELU}
    \end{subfigure}
    \begin{subfigure}{.24\linewidth}
        \centering
        \includegraphics[height=5cm]{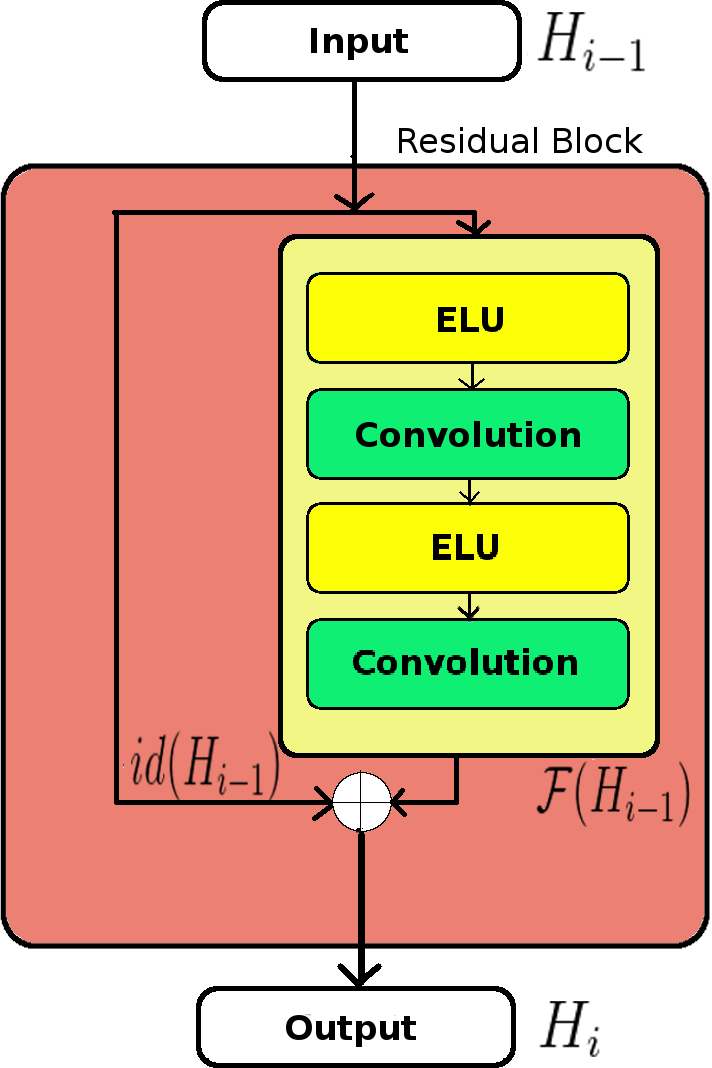}
        \caption{ELU-Conv-ELU-Conv}
        \label{fig:ELU-Conv-ELU-Conv}
    \end{subfigure}
    \begin{subfigure}{.24\linewidth}
        \centering
        \includegraphics[height=5cm]{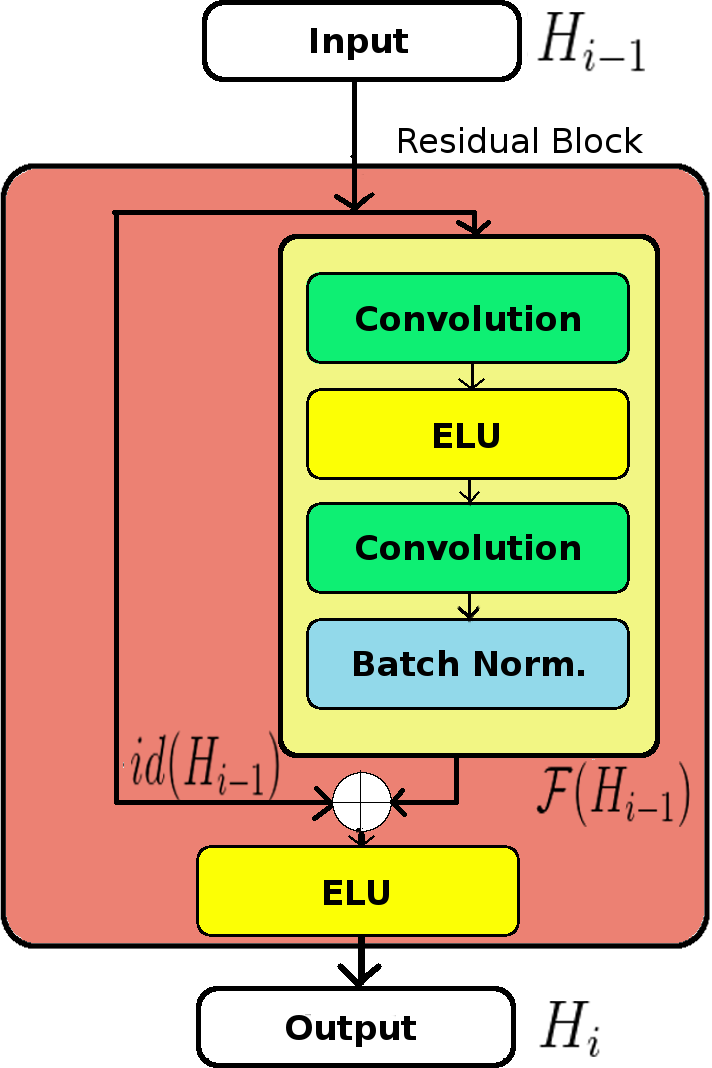}
        \caption{Conv-ELU-Conv-BN and \\ ELU after Addition}
        \label{fig:Conv-ELU-Conv-BN1}
    \end{subfigure}
    \begin{subfigure}{.24\linewidth}
        \centering
        \includegraphics[height=5cm]{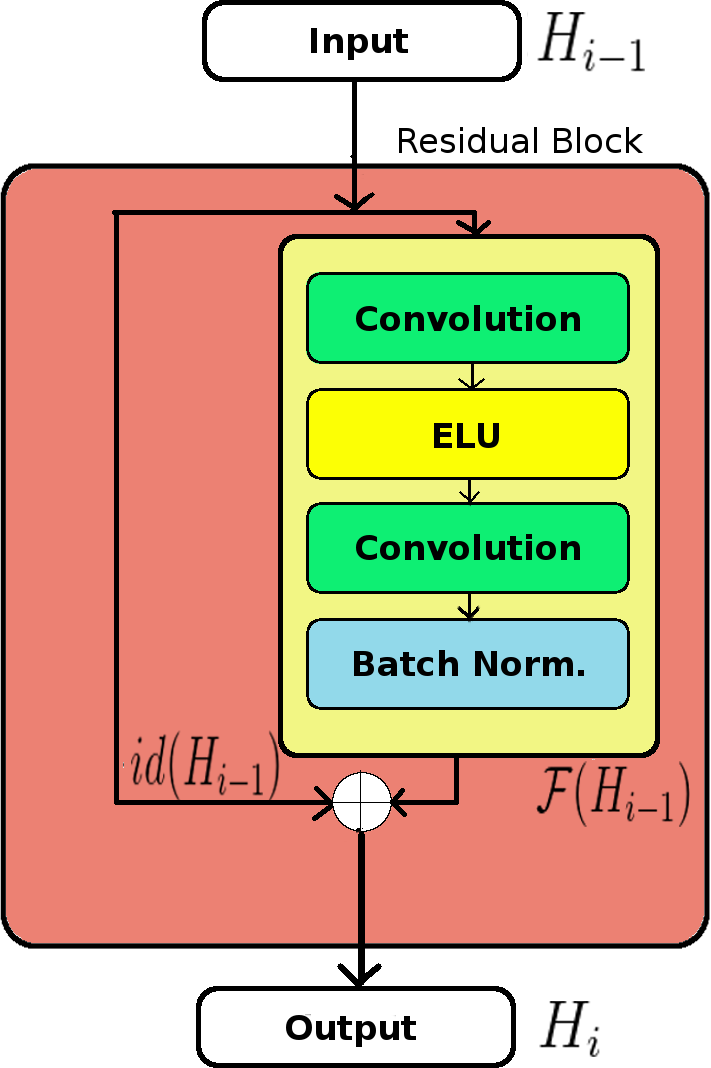}
        \caption{Conv-ELU-Conv-BN and \\ No ELU after Addition}
        \label{fig:Conv-ELU-Conv-BN2}
    \end{subfigure}
    \caption{An \(i^{th}\) Residual Block with Exponential Linear Unit (ELU) in Residual Networks.}
    \label{fig:ELUBlocks}
\end{figure*}

\section{Residual Networks with Exponential Linear Unit}
\subsection{ResNet Architecture}

The Residual Network in \cite{[1]} is a functional composition of \(L\) \emph{residual blocks} (ResBlocks), each encoding the update rule (\ref{eq:1}). Fig \ref{fig:resblock} shows the schematic illustration of the \(i^{th}\) ResBlock. In this example, \(\mathcal{F}\) consists of a sequence of layers: \textbf{Conv-BN-ReLU-Conv-BN}, where Conv and BN stands for Convolution and Batch Normalization respectively. This construction scheme is adopted in all our experiments while reproducing the results of \cite{[1]}. The function \(\mathcal{F}\) is parameterized by some set of  parameters \(W_i\), which we omit for notational simplicity. Normally, we use  64, 32 or 16 filters in the convolutional layers. The size of receptive field is $3 \times 3$. Although it does not seem attractive but, in practice it gives better accuracy without adding any overhead costs, as compared to plain networks.

\subsection{ResNet with ELU}

In comparison with the ResNet model \cite{[1]}, we use Exponential Linear Unit (ELU) in place of a combination of ReLU with Batch Normalization. 
Fig. \ref{fig:ELUBlocks} illustrates our different experiments with ELUs in ResBlock. 

\subsubsection{Conv-ELU-Conv-ELU}

In this model, $\mathcal{F}$ consists of a sequence of layers: \textbf{Conv-ELU-Conv-ELU}. Fig. \ref{fig:Conv-ELU-Conv-ELU} represents the basic building block of this experiment. We trained our model using the specification mentioned in \ref{cifar10analysis}. But we found that after few iterations, the gradients blew up. When the learning rate is decreased, the 20-layer model starts converging but to very less accuracy. The deeper models like 56 and 110-layer still do not converge after decreasing the learning rate. This model clearly fails as the trivial problem of exploding gradient can not be reduced in very deep models.

\subsubsection{ELU-Conv-ELU-Conv}

This is a \textbf{full pre-activation unit} ResBlock \cite{[32]} with ELU. The sequence of layers is \textbf{ELU-Conv-ELU-Conv}. Fig. \ref{fig:ELU-Conv-ELU-Conv} highlights the basic ResBlock of this experiment. During the training of this model too, the gradients exploded after few iterations. Due to the exponential function, the gradients get larger and lead to exploding gradient problem. Even decreasing the learning rate also does not reduce this problem. We decided to add a Batch Normalization layer before Addition to control this problem.

\subsubsection{Conv-ELU-Conv-BN and ELU after Addition}

\begin{figure}
    \centering
    \begin{subfigure}{.23\textwidth}
        \centering
        \includegraphics[width=4cm]{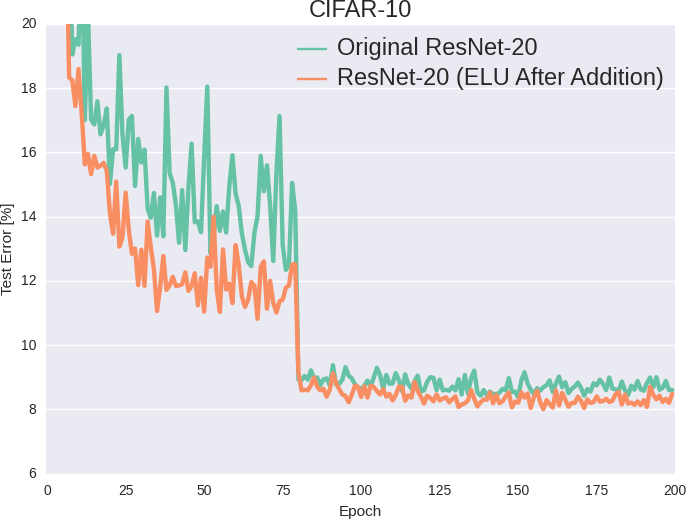}
        \caption{20-layers}
    \end{subfigure}
    \begin{subfigure}{.23\textwidth}
        \centering
        \includegraphics[width=4cm]{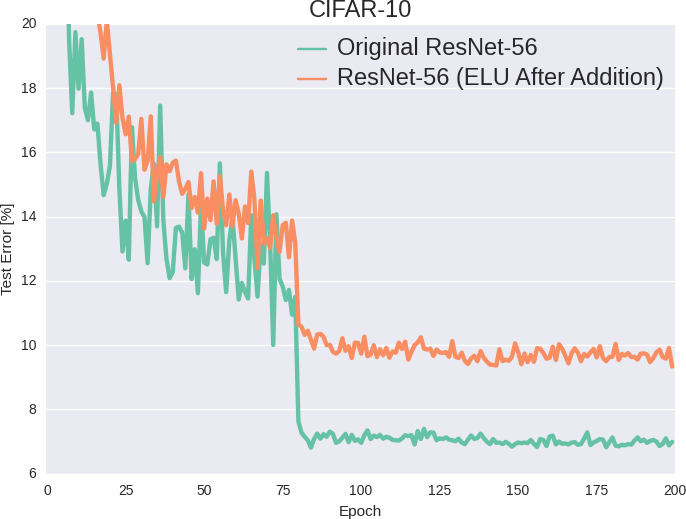}
        \caption{56-layers}
    \end{subfigure}
    \caption{Comparison of test error on CIFAR-10 for the original ResNet model and when ELU is placed after addition in our model.}
    \label{fig:ELU_After_Add}
\end{figure}

To control the exploding gradient, we added a Batch Normalization before addition. So, the sequence of layers in this ResBlock is \textbf{Conv-ELU-Conv-BN} and ELU after addition. Fig. \ref{fig:Conv-ELU-Conv-BN1} represents the ResBlock used in this experiment. Thus in this ResBlock, the update rule (\ref{eq:1}) for the \(i^{th}\) layer is 

\begin{equation} \label{eq:elu_after_add}
 H_i = ELU( \mathcal{F}(H_{i-1}) + id(H_{i-1} ))
\end{equation}

The Batch Normalization layer reduces the exploding gradient problem found in the previous two models. We found that this model gives better accuracy for 20-layer model. However, as we increased the depth of the network, the accuracy degrades for the deeper models. If the ELU activation function is placed after addtion, then the mean activation of the output pushes towards zero. This could be beneficial. However, this forces each skip connection to perturb the output. This has a harmful effect and we found that this leads to degradation of accuracy in very deep ResNets. Fig. \ref{fig:ELU_After_Add} depicts the effects of including ELU after addition in this ResBlock.

\subsubsection{Conv-ELU-Conv-BN and No ELU after Addition}

\begin{figure*}
    \centering
    \begin{subfigure}{.33\linewidth}
        \centering
        \includegraphics[width=5cm]{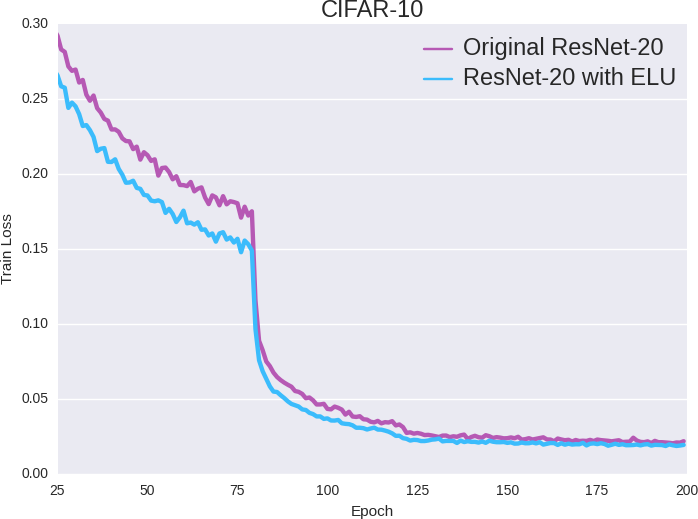}
        \caption{20-layers}
        \label{fig:Cifar10TrainLoss20}
    \end{subfigure}
    \begin{subfigure}{.33\linewidth}
        \centering
        \includegraphics[width=5cm]{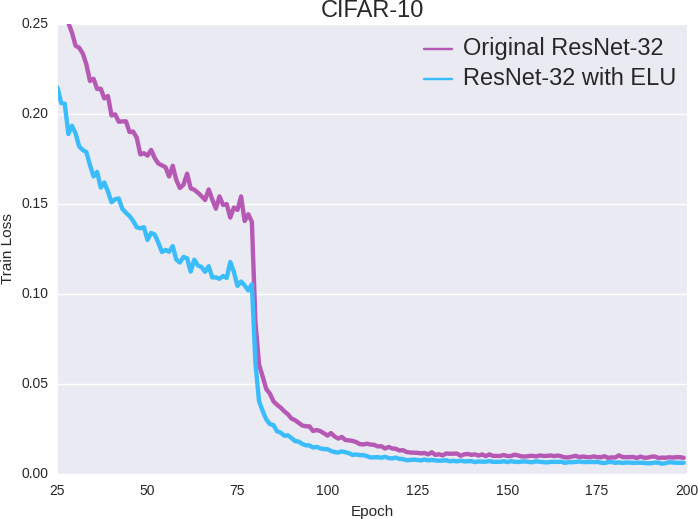}
        \caption{32-layers}
        \label{fig:Cifar10TrainLoss32}
    \end{subfigure}
    \begin{subfigure}{.33\linewidth}
        \centering
        \includegraphics[width=5cm]{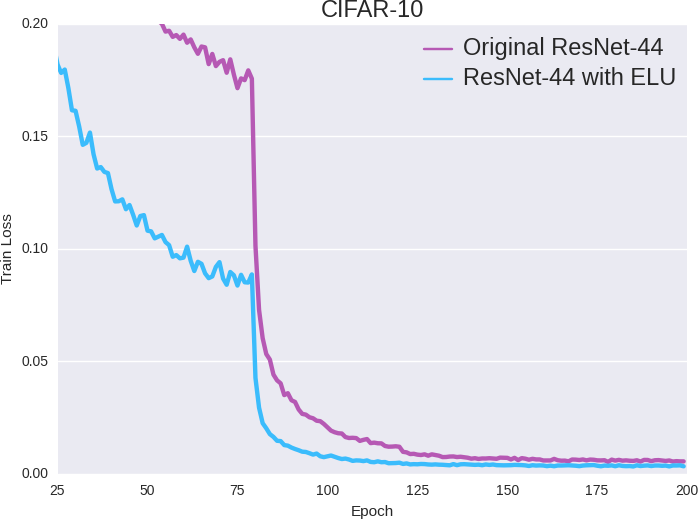}
        \caption{44-layers}
        \label{fig:Cifar10TrainLoss44}
    \end{subfigure}
    \begin{subfigure}{.4\linewidth}
        \centering
        \includegraphics[width=5cm]{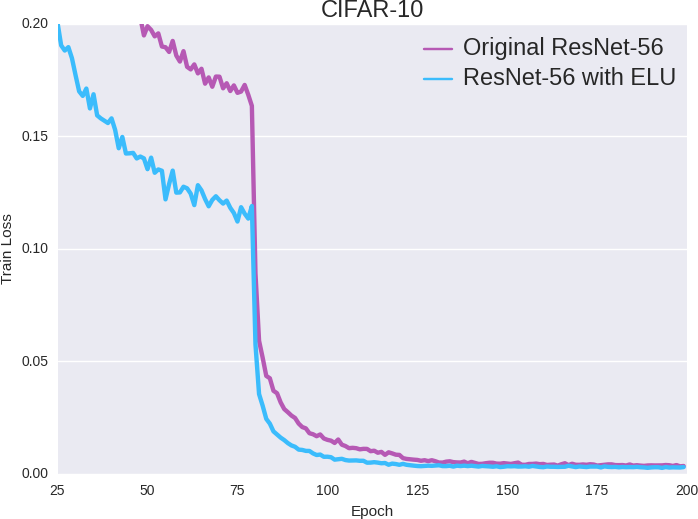}
        \caption{56-layers}
        \label{fig:Cifar10TrainLoss56}
    \end{subfigure}
    \begin{subfigure}{.4\linewidth}
        \centering
        \includegraphics[width=5cm]{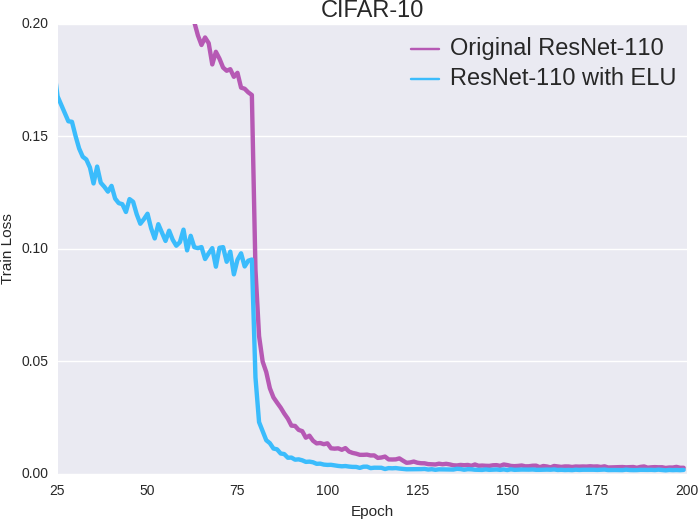}
        \caption{110-layers}
        \label{fig:Cifar10TrainLoss110}
    \end{subfigure}
    \caption{Comparison of the learning behavior of our model and the original ResNet model on CIFAR-10 dataset. We compare the results for 20, 32, 44, 56 and 110-layers and show that our model significantly outperforms the original ResNet model.}
    \label{fig:Cifar10TrainLoss}
\end{figure*}

Fig. \ref{fig:Conv-ELU-Conv-BN2} gives an illustration of the basic building block of our model. Thus in our model, \(\mathcal{F}\) represents the following sequence of layers: \textbf{Conv-ELU-Conv-BN}. The update rule (\ref{eq:1}) for the \(i^{th}\) layer is 

\begin{equation} \label{eq:2}
 H_i = \mathcal{F}(H_{i-1}) + id(H_{i-1})
\end{equation}

This is the basic building block for all our experiments on CIFAR-10 and CIFAR-100 datasets. We show that not including ELU after addition does not degrade the accuracy, unlike the previous model. This ResBlock improves the learning behavior and the classification performance of the Residual Network.


\section{Results}

We empirically demonstrate the effectiveness of our model on a series of benchmark data sets: CIFAR-10 and CIFAR-100. In our experiments, we compare the learning behavior and the classification performance of both the models on the CIFAR-10 and CIFAR-100 datasets. The experiments prove that our model outperforms the original ResNet model in terms of learning behavior and classification performance on both the datasets. Finally, we compare the classification performance of our model with other previously published state-of-the-art models.  

\subsection{CIFAR-10 Analysis}
\label{cifar10analysis}

The first experiment was performed on the CIFAR-10 dataset \cite{[10]}, which consists of 50k training images and 10k test images in 10 classes. In our experiments, we performed training on the training set and evaluation on the test set.

The inputs to the network are $32 \times 32$ images which are color-normalized. We use a $3 \times 3$ receptive field in the convolution layer. We use a stack of $6n$ layers with $3 \times 3$ convolution on the feature maps of sizes $\{32, 16, 8\}$ respectively, with $2n$ on each feature map. The number of filters are $\{16, 32, 64\}$ respectively. The original ResNet model ends with a global average pooling, a 10-way fully-connected layer and a softmax layer. In our model, we add an ELU activation function just before the global average pooling layer.

These two models are trained on a AWS g2.2xlarge instance (which has a single GPU) with a mini batch-size of 128. We use a weight decay of 0.0001 and a momentum of 0.9, and adopt the weight initialization in \cite{[5]} and BN \cite{[4]} but with no dropout. We start with a learning rate of 0.1 and divide by 10 after 81 epochs, and again divide by 10 after 122 epochs. We use the data augmentation mentioned in \cite{[18]} during the training phase: Add 4 pixels on each side and do a random $32 \times 32$ crop from the padded image or its horizontal flip. During the testing phase, we only use a color-normalized $32 \times 32$ image. Our experiments are executed on 20, 32, 44, 56 and 110-layer networks.

\subsubsection{Learning Behavior}

Fig. \ref{fig:Cifar10TrainLoss} shows the comparison of learning behaviours between our model and the original ResNet model on CIFAR-10 dataset for 20, 32, 44, 56 and 110-layers. The graphs prove that for all the different number of layers, our model possesses a superior learning behavior and converges many epochs before the original model. As the depth of the model increases, our model also learns faster than the original model. The difference between the learning rate of these two models increases as the depth increases. Comparing Fig. \ref{fig:Cifar10TrainLoss20} and Fig. \ref{fig:Cifar10TrainLoss110}, we can easily notice the huge difference in learning rates for 20-layer and 110-layer models. After 125 epochs, both the models converge to almost the same value. But, our model has a slightly lower training loss compared to the original model.

\subsubsection{Classification Performance}

\begin{figure*}
    \centering
    \begin{subfigure}{.33\linewidth}
        \centering
        \includegraphics[width=5cm]{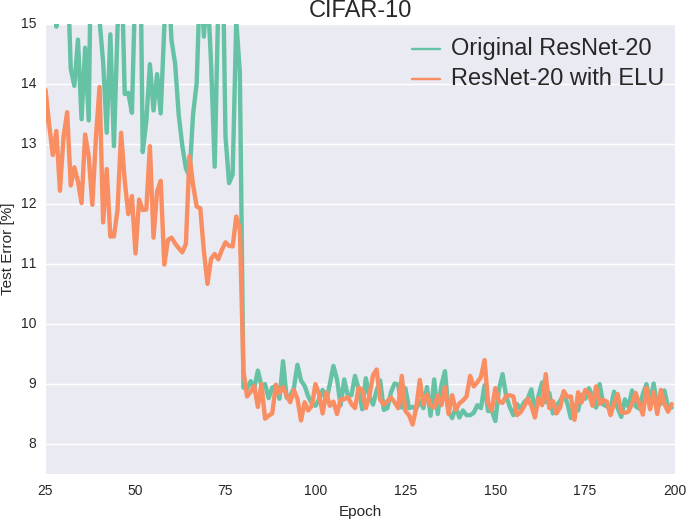}
        \caption{20-layers}
        \label{fig:Cifar10TestError20}
    \end{subfigure}
    \begin{subfigure}{.33\linewidth}
        \centering
        \includegraphics[width=5cm]{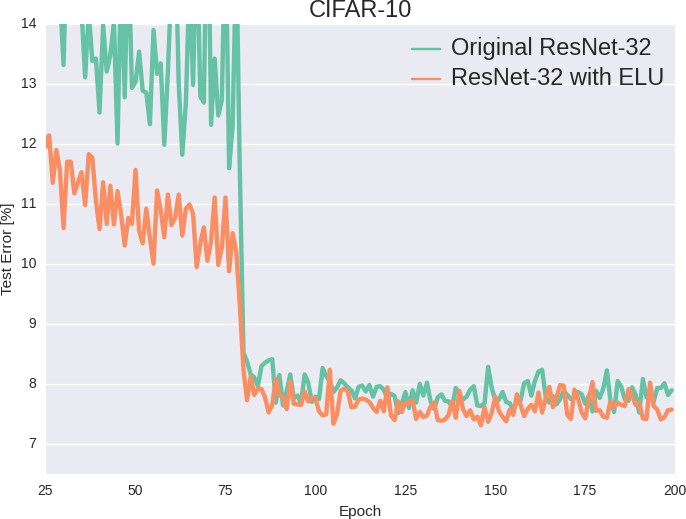}
        \caption{32-layers}
        \label{fig:Cifar10TestError32}
    \end{subfigure}
    \begin{subfigure}{.33\linewidth}
        \centering
        \includegraphics[width=5cm]{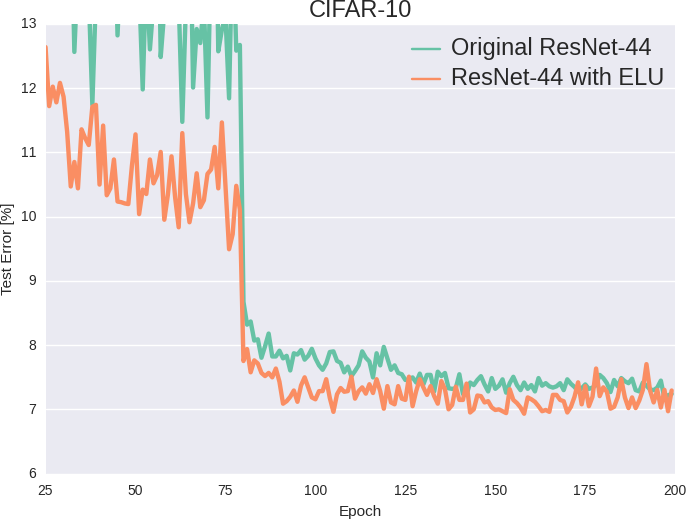}
        \caption{44-layers}
        \label{fig:Cifar10TestError44}
    \end{subfigure}
    \begin{subfigure}{.4\textwidth}
        \centering
        \includegraphics[width=5cm]{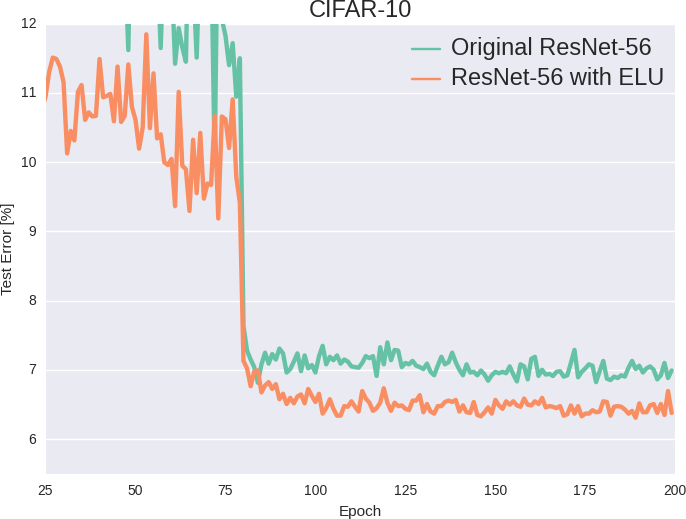}
        \caption{56-layers}
        \label{fig:Cifar10TestError56}
    \end{subfigure}
    \begin{subfigure}{.4\textwidth}
        \centering
        \includegraphics[width=5cm]{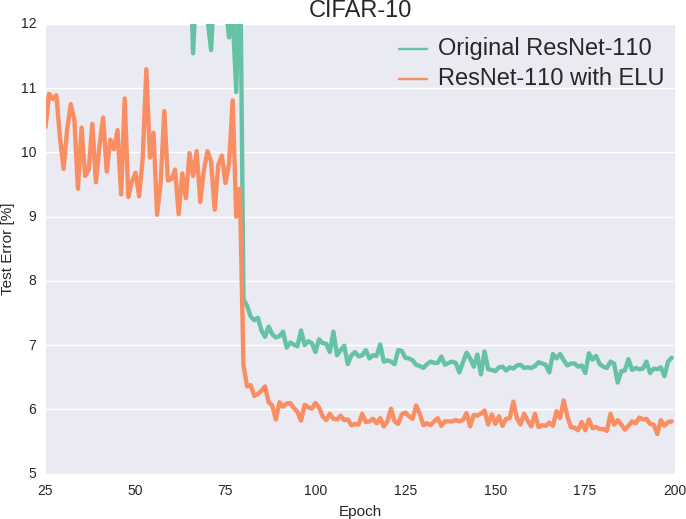}
        \caption{110-layers}
        \label{fig:Cifar10TestError110}
    \end{subfigure}
    \caption{Comparison of the classification performance of our model and the original ResNet model for 20, 32, 44, 56, and 110-layers. We observe that our model outperforms the original one.}
    \label{fig:Cifar10TestError}
\end{figure*}

Fig. \ref{fig:Cifar10TestError} illustrates the comparison of classification performance between our model and the original one on CIFAR-10 dataset for 20, 32, 44, 56 and 110 layers. We observe that for the 20-layer model, the test error is nearly the same for both the models. But, as the depth increases, our model significantly outperforms the original model. Table \ref{table:Cifar10TestErrorTable} shows the test error for both the models from the epoch with the lowest validation error. Fig. \ref{fig:Cifar10TestError} shows that the gap between the test error of the two models increases as the depth is also increased.

\setlength{\tabcolsep}{4pt}
\begin{table}
    \centering
    \caption{Test error (\%) of our model compared to the original ResNet model. The test error of the original ResNet model refers to our reproduction of the experiments by  He et al. \cite{[1]}}
    \label{table:Cifar10TestErrorTable}
    \begin{tabular}{c c c}
        \hline \noalign{\smallskip}
        Layers & Original ResNet & ResNet with ELU \\
        \noalign{\smallskip}
        \hline
        \noalign{\smallskip}
        20 & 8.38 & \textbf{8.32}\\
        32 & 7.51 & \textbf{7.30}\\
        44 & 7.17 & \textbf{6.93}\\
        56 & 6.97 & \textbf{6.31}\\
        110 & 6.42 & \textbf{5.62}\\
        \hline
    \end{tabular}
\end{table}

\subsection{CIFAR-100 Analysis}

\begin{figure}
    \centering
    \begin{subfigure}{.23\textwidth}
        \centering
        \includegraphics[width=4cm]{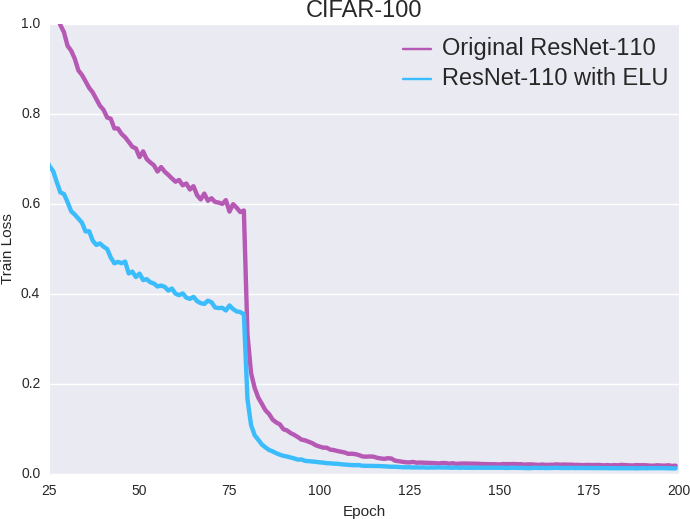}
        \caption{Learning Behavior}
        \label{fig:Cifar100TrainLoss}
    \end{subfigure}
    \begin{subfigure}{.23\textwidth}
        \centering
        \includegraphics[width=4cm]{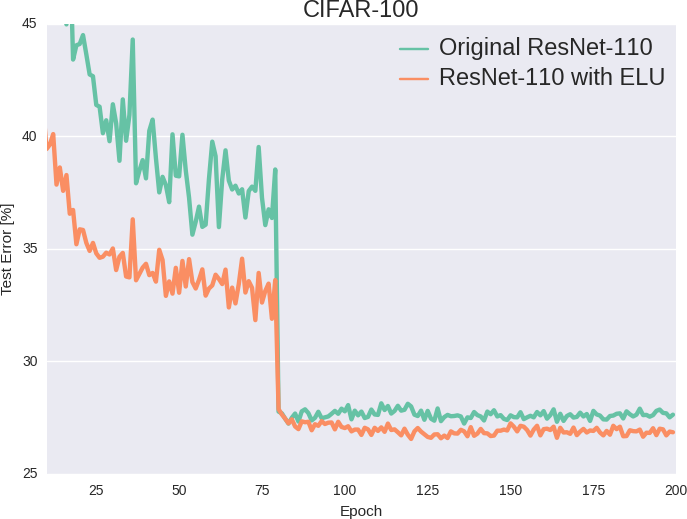}
        \caption{Classification Performance}
        \label{fig:Cifar100TestError}
    \end{subfigure}
    \caption{Comparison of our model and the original ResNet model on CIFAR-100 dataset. We show that our model has superior learning behavior and classification performance compared to the original ResNet model.}
    \label{fig:Cifar100}
\end{figure}

Similar to CIFAR-10, the CIFAR-100 dataset \cite{[10]} also contains $32 \times 32$ images with the same train-test split, but from 100 classes. For both the original model and our model, the experimental settings are exactly the same as those of CIFAR-10. We trained only for the 110-layer models as it gives us state-of-the-art results. Fig. \ref{fig:Cifar100TrainLoss} shows that for CIFAR-100 dataset as well, our model learns faster than the original ResNet model. The original model yields a test error of \textbf{27.23\%}, which is already state-of-the-art in CIFAR-100 with standard data augmentation. Our model reduces the test error to \textbf{26.55\%} and is again one of the best published single model performances. Fig. \ref{fig:Cifar100TestError} shows that the test error of our model is much lower from the starting epoch itself. Table \ref{table:CifarDatasetCompare} shows the comparison of our result with other previously published results on the CIFAR-10 and CIFAR-100 datasets.

\setlength{\tabcolsep}{4pt}
\begin{table}
    \centering
    \caption{Test error (\%) of our model compared to other most competitive methods previously published. All these methods apply standard data augmentation on CIFAR-10 and CIFAR-100 datasets.}
    \label{table:CifarDatasetCompare}
    \begin{tabular}{c c c}
        \hline \noalign{\smallskip}
         & CIFAR-10 & CIFAR-100 \\
        \noalign{\smallskip}
        \hline
        \noalign{\smallskip}
        Maxout \cite{[20]}               &   9.38     &  -    \\
        DropConnect \cite{[19]}          &   9.32     &  -    \\
        Net in Net  \cite{[21]}          &   8.81     &  -    \\
        Deeply Supervised \cite{[18]}    &   7.97     &  -    \\
        Frac. Pool \cite{[22]}           &    -       & 27.62 \\
        All-CNN \cite{[16]}              &   7.25     &  -    \\
        Learning-Activation \cite{[23]}  &   7.51     & 30.83 \\
        R-CNN \cite{[24]}                &   7.09     &  -    \\
        Scalable BO \cite{[25]}          &   6.37     & 27.40 \\
        Highway Networks \cite{[26]}     &   7.60     & 32.24 \\
        Gen. Pool \cite{[27]}            &   6.05     &  -    \\
        \noalign{\smallskip}
        \hline
        ResNet \cite{[1]}               &   6.42     & 27.23 \\
        ResNet with ELU       &   \textbf{5.62}     & \textbf{26.55} \\
        \hline
    \end{tabular}
\end{table}

\section{Conclusions}

In this paper, we introduce Residual Networks with exponential linear units which learn faster than the current Residual Networks. They also give better accuracy than the original ones when the depth is increased. On datasets like CIFAR-10 and CIFAR-100, we improve beyond the current state-of-the-art in terms of test error, while also learning faster than these models using ELUs. ELUs push the mean activations towards zero as they introduce small negative values. This reduces the bias shift and increases the learning speed. Our experiments show that not only does our model have superior learning behavior, but it also provides better accuracy as compared to the current model on CIFAR-10 and CIFAR-100 datasets. This enables the researchers to use very deep models and also increase their learning behavior and classification performance at the same time.

{\small
\bibliographystyle{ieee}

}

\end{document}